%% file: acl2021.tex
\pdfoutput=1

\documentclass[11pt,a4paper]{article}

\usepackage[hyperref]{acl2021}

\usepackage{times}
\usepackage{latexsym}
\usepackage{soul}
\usepackage{enumitem}
\usepackage{amssymb}
\usepackage{parskip}
\usepackage{booktabs}
\usepackage[T1]{fontenc}

\usepackage[utf8]{inputenc}
\usepackage{graphicx}
\usepackage{multirow}
\usepackage{microtype}

\aclfinalcopy
\usepackage{algorithmicx}
\usepackage{algpseudocode}
\usepackage{algorithm}
\usepackage{amsmath}
\usepackage[nameinlink]{cleveref}
\input{math_commands}

\definecolor{kmy-color}{rgb}{0.858, 0.188, 0.478}

\title{Reliability Testing for Natural Language Processing Systems}

\author{Samson Tan$^{\S\natural}$\thanks{\; Correspondence to:\fontsize{8.5}{9} \texttt{samson.tan@salesforce.com}} \quad Shafiq Joty$^{\S\ddagger}$ \quad Kathy Baxter$^{\S}$ \\ \textbf{Araz Taeihagh$^{\diamondsuit\clubsuit}$ \quad Gregory A.\ Bennett$^{\S}$ \quad Min-Yen Kan$^\natural$}
\vspace{0.5em}\\
  $^\S$Salesforce Research \\
  $^\ddagger$Nanyang Technological University \\
  $^\natural$School of Computing, National University of Singapore \\
  $^\diamondsuit$Lee Kuan Yew School of Public Policy, National University of Singapore\\
  $^\clubsuit$Centre for Trusted Internet and Community, National University of Singapore
}

\begin{document}
\maketitle
\begin{abstract}
Questions of fairness, robustness, and transparency are paramount to address before deploying NLP systems. Central to these concerns is the question of reliability: Can NLP systems reliably treat different demographics fairly \emph{and} function correctly in diverse and noisy environments? To address this, we argue for the need for reliability testing and contextualize it among existing work on improving accountability. We show how adversarial attacks can be reframed for this goal, via a framework for developing reliability tests. We argue that reliability testing --- with an emphasis on interdisciplinary collaboration --- will enable rigorous and targeted testing, and aid in the enactment and enforcement of industry standards.
\end{abstract}

\section{Introduction}\label{sec:intro}
Rigorous testing is critical to ensuring a program works as intended (functionality) when used under real-world conditions (reliability). Hence, it is troubling that while natural language technologies are becoming increasingly pervasive in our everyday lives, there is little assurance that these NLP systems will not fail catastrophically or amplify discrimination against minority demographics when exposed to input from outside the training distribution. Recent examples include GPT-3 \citep{gpt3} agreeing with suggested suicide \citep{drgpt}, the mistranslation of an innocuous social media post resulting in a minority's arrest \citep{translation-arrest2017}, and biased grading algorithms that can negatively impact a minority student's future \citep{feathersflawed2019}. Additionally, a lack of rigorous testing, coupled with machine learning's (ML) implicit assumption of identical training and testing distributions, may inadvertently result in systems that discriminate against minorities, who are often underrepresented in the training data. This can take the form of misrepresentation of or poorer performance for people with disabilities, specific gender, ethnic, age, or linguistic groups \citep{hovy2016social,crawford2017,hutchinson-etal-2020-social}.

\begin{figure}[t]
    \centering
    \includegraphics[width=0.5\textwidth]{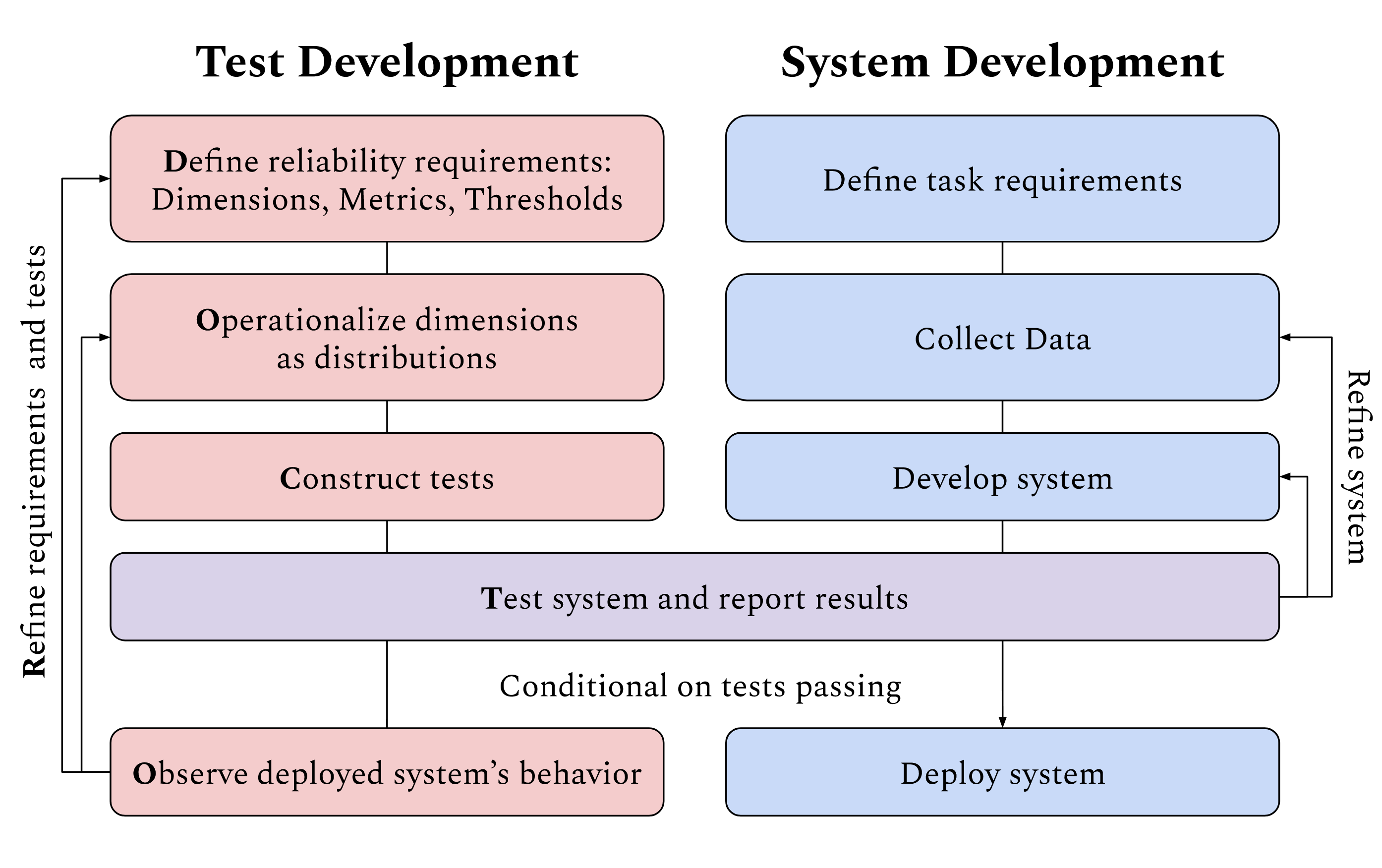}
    \caption{How DOCTOR can integrate with existing system development workflows. Test (left) and system development (right) take place in parallel, separate teams. Reliability tests can thus be constructed independent of the system development team, either by an internal ``red team'' or by independent auditors.
    }
    \label{fig:test-system}
\end{figure}

Amongst claims of NLP systems achieving human parity in challenging tasks such as question answering \citep{wei2018fast}, machine translation \citep{hassan2018achieving}, and commonsense inference \citep{devlin2019bert}, research has demonstrated these systems' fragility to natural and adversarial noise \citep{goodfellow2015,belinkov2018synthetic} and out-of-distribution data \citep{fisch-etal-2019-mrqa}. It is also still common practice to equate ``testing'' with ``measuring held-out accuracy'', even as datasets are revealed to be harmfully biased \citep{wagner2015s,geva-etal-2019-modeling,sap-etal-2019-risk}.

Many potential harms can be mitigated by detecting them early and preventing the offending model from being put into production. Hence, in addition to being mindful of the biases in the NLP pipeline \citep{bender2018data,mitchell2019model,waseem2021disembodied} and holding creators accountable via audits \citep{raji2020closing,brundage2020toward}, we argue for the need to evaluate an NLP system's reliability in diverse operating conditions.

Initial research on evaluating out-of-distribution generalization involved manually-designed challenge sets \citep{jia-liang-2017-adversarial,nie-etal-2020-adversarial,gardner-etal-2020-evaluating}, counterfactuals \citep{kaushik2019learning,khashabi-etal-2020-bang,wu2021polyjuice},  biased sampling \citep{sogaard2021we} or toolkits for testing if a system has specific capabilities \citep{ribeiro-etal-2020-beyond} or robustness to distribution shifts \citep{goel2021robustness}. However, most of these approaches inevitably overestimate a given system's \emph{worst-case performance} since they do not mimic the NLP system's adversarial distribution\footnote{The distribution of adversarial cases or failure profile.}.

A promising technique for evaluating worst-case performance is the adversarial attack. However, although some adversarial attacks explicitly focus on specific linguistic levels of analysis \citep{belinkov2018synthetic,iyyer-etal-2018-adversarial,tan-etal-2020-morphin,eger-benz-2020-hero}, many often simply rely on word embeddings or language models for perturbation proposal (see \Cref{sec:atk}). While the latter may be useful to evaluate a system's robustness to malicious actors, they are less useful for dimension-specific testing (e.g., reliability when encountering grammatical variation). This is because they often perturb the input across multiple dimensions at once, which may make the resulting adversaries unnatural.

Hence, in this paper targeted at NLP researchers, practitioners, and policymakers, we make the case for reliability testing and reformulate adversarial attacks as \emph{dimension-specific}, worst-case tests that can be used to approximate real-world variation. We contribute a reliability testing framework --- DOCTOR ---  that translates safety and fairness concerns around NLP systems into quantitative tests. We demonstrate how testing dimensions for DOCTOR can be drafted for a specific use case. Finally, we discuss the policy implications, challenges, and directions for future research on reliability testing.

\section{Terminology Definitions}
Let's define key terms to be used in our discussion.

\paragraph{NLP system.}
The entire text processing pipeline built to solve a specific task; taking raw text as input and producing predictions in the form of labels (classification) or text (generation). We exclude \emph{raw} language models from the discussion since it is unclear how performance, and hence worst-case performance, should be evaluated. We do include NLP systems that use language models internally (e.g., BERT-based classifiers \citep{devlin2019bert}).

\paragraph{Reliability.}
Defined by \citet{8016712} as the ``degree to which a system, product or component performs specified functions under specified conditions for a specified period of time''.  We prefer this term over robustness\footnote{The ``degree to which a system or component can function correctly in the presence of invalid inputs or stressful environmental conditions'' \citep{8016712}.} to challenge the NLP community's common framing of inputs from outside the training distribution as ``noisy''. The notion of reliability requires us to explicitly consider the specific, diverse environments (i.e., communities) a system will operate in.  This is crucial to reducing the NLP's negative impact on the underrepresented.

\paragraph{Dimension.} An axis along which variation can occur in the real world, similar to \citet{planknon2016}'s variety space. A taxonomy of possible dimensions can be found in \Cref{tab:taxonomy} (Appendix).

\paragraph{Adversarial attack.}
A method of perturbing the input to degrade a target model's accuracy \citep{goodfellow2015}. In computer vision, this is achieved by adding adversarial noise to the image, optimized to be maximally damaging to the model. \Cref{sec:atk} describes how this is done in the NLP context.

\paragraph{Stakeholder.}
A person who is (in-)directly impacted by the NLP system's predictions.

\paragraph{Actor.}
Someone who has influence over a) the design of an NLP system and its reliability testing regime; b) whether the system is deployed; and c) who it can interact with. Within the context of our discussion, actors are likely to be regulators, experts, and stakeholder advocates.

\paragraph{\hspace{0.75em} {\it \textbf{Expert.}}}
An actor who has specialized knowledge, such as ethicists, linguists, domain experts, social scientists, or NLP practitioners.

\section{The Case for Reliability Testing in NLP}\label{sec:doesnlp}
The accelerating interest in building NLP-based products that impact many lives has led to urgent questions of fairness, safety, and accountability \citep{hovy2016social,bender2021}, prompting research into algorithmic bias  \citep{bolukbasi2016,blodgett-etal-2020-language}, explainability \citep{lime2016,danilevsky-etal-2020-survey}, robustness \citep{jia-liang-2017-adversarial}, etc. Research is also emerging on best practices for productizing ML: from detailed dataset documentation  \citep{bender2018data,gebru2018datasheets}, model documentation for highlighting important but often unreported details such as its training data, intended use, and caveats \citep{mitchell2019model}, and documentation best practices  \citep{aboutml2019}, to institutional mechanisms such as auditing \citep{raji2020closing} to enforce accountability and red-teaming \citep{brundage2020toward} to address developer blind spots, not to mention studies on the impact of organizational structures on responsible AI initiatives \citep{rakova2020responsible}.

Calls for increased accountability and transparency are gaining traction among governments \citep{aaa2019,nistai2019,euaiwhitepaper2020,ftc2020,cpra2020,fda2021} and customers increasingly cite ethical concerns as a reason for not engaging AI service providers \citep{eiu2020}.

While there has been significant discussion around best practices for dataset and model creation, work to ensure NLP systems are evaluated in a manner representative of their operational conditions has only just begun. Initial work in constructing representative tests focuses on enabling development teams to easily evaluate their models' linguistic capabilities \cite{ribeiro-etal-2020-beyond} and accuracy on subpopulations and distribution shifts \citep{goel2021robustness}. However, there is a clear need for a paradigm that allows experts and stakeholder advocates to collaboratively develop tests that are representative of the practical and ethical concerns of an NLP system's target demographic. We argue that \emph{reliability testing}, by reframing the concept of adversarial attacks, has the potential to fill this gap.

\subsection{What is reliability testing?}
Despite the recent advances in neural architectures resulting in breakthrough performance on benchmark datasets, research into adversarial examples and  out-of-distribution generalization has found ML systems to be particularly vulnerable to slight perturbations in the input \citep{goodfellow2015} and natural distribution shifts \citep{fisch-etal-2019-mrqa}. While these perturbations are often chosen to maximize model failure, they highlight serious reliability issues for putting ML models into production since they show that these models \emph{could} fail catastrophically in naturally noisy, diverse, real-world environments \citep{saria2019}. Additionally, bias can seep into the system at multiple stages of the NLP lifecycle \citep{shah-etal-2020-predictive}, resulting in discrimination against minority groups \citep{o2016weapons}. The good news, however, is that rigorous testing can help to highlight potential issues before the systems are deployed.

The need for rigorous testing in NLP is reflected in ACL 2020 giving the Best Paper Award to CheckList \citep{ribeiro-etal-2020-beyond}, which applied the idea of behavior testing from software engineering to testing NLP systems. While invaluable as a first step towards the development of comprehensive testing methodology, the current implementation of CheckList may still overestimate the reliability of NLP systems since the individual test examples are largely manually constructed. Importantly, with the complexity and scale of current models, humans cannot accurately determine a model's \emph{adversarial} distribution
(i.e., the examples that cause model failure). Consequently, the test examples they construct are unlikely to be the \emph{worst-case examples} for the model. Automated assistance is needed.

Therefore, we propose to perform \emph{reliability testing}, which can be thought of as one component of behavior testing. We categorize reliability tests as \emph{average-case} tests or the \emph{worst-case} tests. As their names suggest, average-case and worst-case tests estimate the expected and lower-bound performance, respectively, when the NLP system is exposed to the phenomena modeled by the tests. Average-case tests are conceptually similar to \citet{wu2021polyjuice}'s counterfactuals, which is contemporaneous work, while worst-case tests are most similar to adversarial attacks (\Cref{sec:atk}).

Our approach parallels \emph{boundary value testing} in software engineering: In boundary value testing, tests evaluate a program's ability to handle edge cases using test examples drawn from the extremes of the ranges the program is expected to handle. Similarly, reliability testing aims to quantify the system's reliability under diverse and potentially extreme conditions. This allows teams to perform better quality control of their NLP systems and introduce more nuance into discussions of why and when models fail (\Cref{sec:framework}).
Finally, we note that reliability testing and standards are established practices in engineering industries (e.g., aerospace \citep{aerospaceproc2003,adaptiveverification2016}) and advocate for NL engineering to be at parity with these fields.

\subsection{Evaluating worst-case performance in a label-scarce world}
A proposed approach for testing robustness to natural and adverse distribution shifts is to construct test sets using data from different domains or writing styles \citep{miller2020effect,hendrycks-etal-2020-pretrained}, or to use a human vs.\ model method of constructing challenge sets \citep{nie-etal-2020-adversarial,zhang-etal-2019-paws}. While they are the gold standard, such datasets are expensive to construct,\footnote{\citet{dua-etal-2019-drop} reports a cost of 60k USD for 96k question--answer pairs.} making it infeasible to manually create worst-case test examples for each NLP system being evaluated. Consequently, these challenge sets necessarily overestimate each system's worst-case performance when the inference distribution differs from the training one. Additionally, due to their crowdsourced nature, these challenge sets inevitably introduce distribution shifts across multiple dimensions at once, and even their own biases \citep{geva-etal-2019-modeling}, unless explicitly controlled for. Building individual challenge sets for each dimension would be prohibitively expensive due to combinatorial explosion, even before having to account for concept drift \citep{Widmer1996LearningIT}. This coupling complicates efforts to design a nuanced and comprehensive testing regime. Hence, simulating variation in a controlled manner via reliability tests can be a complementary method of evaluating the system's out-of-distribution generalization ability.

\section{Adversarial Attacks as Reliability Tests}\label{sec:atk}
We first give a brief introduction to adversarial attacks in NLP before showing how they can be used for reliability testing. We refer the reader to \citet{zhangADVsurvey} for a comprehensive survey.

\begin{algorithm}[t]
\small
\begin{algorithmic}[1]
\Require Data distribution $\gD_d = \{\gX, \gY\}$ modeling the dimension of interest $d$, NLP system $\gM$, Source dataset $X \sim \gX$, Desired labels $Y'\sim \gY$, Scoring function $S$.
\Ensure Average- or worst-case examples $X'$, Result $r$.

\State $X' \gets \{\varnothing\}$, $r \gets 0$

\For{$x, y'$ in $X, Y'$}
\State $C \gets$  \Call{SampleCandidates}{$\gX$}

\Switch{TestType}
\Case{AverageCaseTest}
\State $s \gets \Call{Mean}{\gS(y', \gM(C))}$
\State $X' \gets X' \cup C$
\EndCase

\Case{WorstCaseTest}
\State $x', s \gets \argmin_{x_c \in C} \; \gS(y',\gM(x_c))$
\State $X' \gets X' \cup \{x'\}$
\EndCase
\EndSwitch

\State $r \gets r + s$
\EndFor

\State $r \gets \frac{r}{|X|}$
\State \Return $X', r$
\end{algorithmic}
\caption{General Reliability Test}
\label{algo:test}
\end{algorithm}

Existing work on NLP adversarial attacks perturbs the input at various levels of linguistic analysis: phonology \citep{eger-benz-2020-hero}, orthography \citep{ebrahimi-etal-2018-hotflip}, morphology \citep{tan-etal-2020-morphin}, lexicon \citep{alzantot-etal-2018-generating,jin2019bert}, and syntax \citep{iyyer-etal-2018-adversarial}.

Early work did not place any constraints on the attacks and merely used the degradation to a target model's accuracy as \emph{the} measure of success. However, this often resulted in the semantics and expected prediction changing, leading to an over-estimation of the attack's success. Recent attacks aim to preserve the original input's semantics. A popular approach has been to substitute words with their synonyms using word embeddings or a language model as a measure of semantic similarity \citep{alzantot-etal-2018-generating,SinghGR18,michel-etal-2019-evaluation,ren2019generating,zhang-etal-2019-generating,li2019textbugger,jin2019bert,garg2020bae,li2020bert}.

Focusing on maximally degrading model accuracy overlooks the key feature of adversarial attacks: the ability to find the worst-case example for a model from an arbitrary distribution. Many recent attacks perturb the input across multiple dimensions at once, which may make the result unnatural. By constraining our sample perturbations to a distribution modeling a specific dimension of interest, the performance on the generated adversaries is a valid lower bound performance for that dimension. Said another way, adversarial attacks can be reframed as interpretable reliability tests if we constrain them to meaningful distributions.

This is the key element of our approach as detailed in \Cref{algo:test}. We specify either an average (Lines~5--7) or worse case test (Lines~8--10), but conditioned on the data distribution $\gD$ that models a particular dimension of interest $d$. The resultant reliability score gauges real-world performance and the worst-case variant returns the adversarial examples that cause worst-case performance.  When invariance to input variation is expected, $y'$ is equivalent to the source label $y$. Note that by ignoring the average-case test logic and removing $d$, we recover the general adversarial attack algorithm.

However, the key difference between an adversarial robustness mindset and a testing one is the latter's emphasis on identifying ways in which natural phenomena or ethical concerns can be operationalized as reliability tests. This change in perspective opens up new avenues for interdisciplinary research that will allow researchers and practitioners to have more nuanced discussions about model reliability and can be used to design comprehensive reliability testing regimes. We describe such a framework for interdisciplinary collaboration next.

\section{A Framework for Reliability Testing}\label{sec:framework}

We introduce and then describe our general framework, DOCTOR, for testing the reliability of NLP systems. DOCTOR comprises six steps:
\begin{enumerate}[leftmargin=*]
\itemsep-0.3em
    \item \textbf{D}efine reliability requirements
    \item \textbf{O}perationalize dimensions
as distributions
    \item \textbf{C}onstruct tests
    \item \textbf{T}est system and report results
    \item \textbf{O}bserve deployed system's behavior
    \item \textbf{R}efine reliability requirements and tests
\end{enumerate}

\paragraph{Defining reliability requirements.}
Before any tests are constructed, experts and stakeholder advocates should work together to understand the demographics and values of the communities the NLP system will interact with \citep{friedman2019value} and the system's impact on their lives. The latter is also known as algorithmic risk assessment \citep{blackboxLovelace2021}. There are three critical questions
to address:
1)~Along what dimensions should the model be tested? 2)~What metrics should be used to measure system performance? 3)~What are acceptable performance thresholds for each dimension?

Question~1 can be further broken down into: a)~general linguistic phenomena, such as alternative spellings or code-mixing; b)~task-specific quirks, e.g., an essay grading system should not use text length to predict score; c)~sensitive attributes, such as gender, ethnicity, sexual orientation, age, or disability status. This presents an opportunity for interdisciplinary expert collaboration: Linguists are best equipped to contribute to discussions around~(a), domain experts to~(b), and ethicists and social scientists to~(c). However, we recognize that such collaboration may not be feasible for \emph{every} NLP system being tested. It is more realistic to expect ethicists to be involved when applying DOCTOR at the  company and industry levels, and ethics-trained NLP practitioners to answer these questions within the development team. We provide a taxonomy of potential dimensions in \Cref{tab:taxonomy} (Appendix).

Since it is likely unfeasible to test every possible dimension, stakeholder advocates should be involved to ensure their values and interests are accurately represented and prioritized \citep{hagerty2019global}, while experts should ensure the dimensions identified can be feasibly tested. A similar approach to that of community juries\footnote{\href{https://docs.microsoft.com/en-us/azure/architecture/guide/responsible-innovation/community-jury}{docs.microsoft.com/en-us/azure/.../community-jury}} may be taken. We recommend using this question to evaluate the feasibility of operationalizing potential dimensions: ``What is the system's performance when exposed to variation along dimension $d$?''. For example, rather than simply ``gender'', a better-defined dimension would be ``gender pronouns''. With this understanding, experts and policymakers can then create a set of \emph{reliability requirements}, comprising the testing dimensions, performance metric(s), and passing thresholds.

Next, we recommend using the same metrics for held-out, average-case, and worst-case performance for easy comparison. These often vary from task to task and are still a subject of active research \citep{novikova-etal-2017-need,reiter-2018-structured,kryscinski-etal-2019-neural}, hence the question of the right metric to use is beyond the scope of this paper. Finally, ethicists, in consultation with the other aforementioned experts and stakeholders, will determine acceptable thresholds for worst-case performance. The system under test must perform above said thresholds when exposed to variation along those dimensions in order to pass. For worst-case performance, we recommend reporting thresholds as relative differences ($\delta$) between the average-case and worst-case performance. These questions may help in applying this step and deciding if specific NLP solutions should even exist \citep{leins-etal-2020-give}:

\begin{itemize}[leftmargin=*]
\itemsep-0.3em
    \item Who will interact with the NLP system, in what context, and using which language varieties?
    \item What are the distinguishing features of these varieties compared to those used for training?
    \item What is the (short- and long-term) impact on the community's most underrepresented members if the system performs more poorly for them?
\end{itemize}

We note that our framework is general enough to be applied at various levels of organization: within the development team, within the company (compliance team, internal auditor), and within the industry (self-regulation or independent regulator). However, we expect the exact set of dimensions, metrics and acceptable thresholds defined in Step~1 to vary depending on the reliability concerns of the actors at each level. For example, independent regulators will be most concerned with establishing minimum safety and fairness standards that all NLP systems used in their industries must meet, while compliance teams may wish to have stricter and more comprehensive standards for brand reasons. Developers can use DOCTOR to meet the other two levels of requirements and understand their system's behaviour better with targeted testing.

\paragraph{Operationalizing dimensions.} While the abstractness of dimensions allows people who are not NLP practitioners to participate in drafting the set of reliability requirements, there is no way to test NLP systems using fuzzy concepts. Therefore, every dimension the system is to be tested along must be operationalizable as a distribution from which perturbed examples can be sampled in order for NLP practitioners to realize them as tests.

Since average-case tests attempt to estimate a system's expected performance in its deployed environment, the availability of datasets that reflect real-world distributions is paramount to ensure that the tests themselves are unbiased. This is less of an issue for worst-case tests; the tests  only needs to know which perturbations that are possible, but not how frequently they occur in the real world. Figuring out key dimensions for different classes of NLP tasks and exploring ways of operationalizing them as reliability tests are also promising directions for future research. Such research would help NLP practitioners and policymakers define reliability requirements that can be feasibly implemented.

\paragraph{Constructing tests.}
Next, average- and worst-case tests are constructed (\Cref{algo:test}). Average-case tests can be data-driven and could take the form of manually curated datasets or model-based perturbation generation (e.g., PolyJuice \citep{wu2021polyjuice}), while worst-case tests can be rule-based (e.g., Morpheus \citep{tan-etal-2020-morphin}) or model-based (e.g., BERT-Attack \citep{li2020bert}). We recommend constructing tests that do not require access to the NLP model's parameters (black-box assumption); this not only yields more system-agnostic tests, but also allows for (some) tests to be created independently from the system development team. If the black-box assumption proves limiting, the community can establish a standard set of items an NLP system should export for testing purposes, e.g., network gradients if the system uses a neural model. Regardless of assumption, keeping the regulators' test implementations separate and hidden from the system developers is critical for stakeholders and regulators to trust the results. This separation also reduces overfitting to the test suite.

\paragraph{Testing systems.} A possible model for test ownership is to have independently implemented tests at the three levels of organization described above (team, company, industry). At the development team level, reliability tests can be used to diagnose weaknesses with the goal of improving the NLP system for a specific use case and set of target users. Compared to unconstrained adversarial examples, contrasting worst-case examples that have been constrained along specific dimensions with non-worst-case examples will likely yield greater intuition into the model's inner workings. Studying how modifications (to the architecture, training data and process) affect the system's reliability on each dimension will also give engineers insight into the factors affecting system reliability. These tests should be executed and updated regularly during development, according to software engineering best practices such as Agile \citep{beck2001agile}.

Red teams are company-internal teams tasked with finding security vulnerabilities in their developed software or systems.
\citet{brundage2020toward} propose to apply the concept of red teaming to surface flaws in an AI system's  safety and security. In companies that maintain multiple NLP systems, we propose employing similar, specialized teams composed of NLP experts to build and maintain reliability tests that ensure their NLP systems adhere to company-level reliability standards. These tests will likely be less task-/domain-specific than those developed by engineering teams due to their wider scope, while the reliability standards may be created and maintained by compliance teams or the red teams themselves. Making these standards available for public scrutiny and ensuring their products meet them will enable companies to build trust with their users. To ensure all NLP systems meet the company's reliability standards, these reliability tests should be executed as a part of regular internal audits \citep{raji2020closing}, investigative audits after incidents, and before major releases (especially if it is the system's first release or if it received a major update). They may also be regularly executed on randomly chosen production systems and trigger an alert upon failure.

At the independent regulator level, reliability tests would likely be carried out during product certification (e.g., ANSI/ISO certification) and external audits. These industry-level reliability standards and tests may be developed in a similar manner to the company-level ones. However, we expect them to be more general and less comprehensive than the latter, analogous to minimum safety standards such as IEC 60335-1 \citep{iec60335-1}. Naturally, high risk applications and NLP systems used in regulated industries should comply with more stringent requirements \citep{euaiact2021}.

Our proposed framework is also highly compatible with the use of model cards \citep{mitchell2019model} for auditing and transparent reporting \citep{raji2020closing}. In addition to performance on task-related metrics, model cards surface information and assumptions about a machine learning system and training process that may not be readily available otherwise. When a system has passed all tests and is ready to be deployed, its average- and worst-case performance on all tested dimensions can be included as an extra section on the accompanying model card. In addition, the perturbed examples generated during testing and their labels ($x'$, $y'$) can be stored for audit purposes or examined to ensure that the tests are performing as expected.

\vspace{-0.1em}
\paragraph{Observing and Refining requirements.}
It is crucial to regularly monitor the systems' impact post-launch and add, update, or re-prioritize dimensions and thresholds accordingly. Monitoring large-scale deployments can be done via community juries, in which stakeholders who will be likely impacted (or their advocates) give feedback on their pain points and raise concerns about potential negative effects. Smaller teams without the resources to organize community juries can set up avenues (e.g., online forms) for affected stakeholders to give feedback, raise concerns, and seek remediation.

\section{From Concerns to Dimensions}\label{sec:case}
We now illustrate how reliability concerns can be converted into concrete testing dimensions (Step~1) by considering the scenario of applying automated text scoring to short answers and essays from students in the multilingual population of Singapore. We study a second scenario in \Cref{app:case}.
{\bf Automated Text Scoring (ATS)} systems are increasingly used to grade tests and essays \citep{markoffGrading2013,feathersflawed2019}. While they can provide instant feedback and help teachers and test agencies cope with large loads, studies have shown that they often exhibit demographic and language biases, such as scoring African- and Indian-American males lower on the GRE Argument task compared to human graders \citep{bridgeman2012comparison,ramineni2018understanding}. Since the results of some tests will affect the futures of the test takers \citep{insidertesting2018}, the scoring algorithms used must be sufficiently reliable. Hence, let us imagine that Singapore's education ministry has decided to create a standard set of reliability requirements that all ATS systems used in education must adhere to.

\paragraph{Linguistic landscape.}
A mix of language varieties are used in Singapore: a prestige English variety, a colloquial English variety, three other official languages (Chinese, Malay, and Tamil), and a large number of other languages. English is the \emph{lingua franca}, with fluency in the prestige variety correlating with socioeconomic status \citep{vaish2008language}. A significant portion of the population does not speak English at home. Subjects other than languages are taught in English.

\vspace{-0.3em}
\paragraph{Stakeholder impact.} The key stakeholders affected by ATS systems would be students in schools and universities. The consequences of lower scores could be life-altering for the student who is unable to enroll in the major of their choice. At the population level, biases in an ATS system trained on normally sampled data would unfairly discriminate against already underrepresented groups. Additionally, biases against disfluent or ungrammatical text when they are not the tested attributes would result in discrimination against students with a lower socioeconomic status or for whom English is a second language.

Finally, NLP systems have also been known to be overly sensitive to alternative spellings \citep{belinkov2018synthetic}. When used to score subject tests, this could result in the ATS system unfairly penalizing dyslexic students \citep{coleman2009comparison}. Since education is often credited with enabling social mobility,\footnote{\href{https://www.encyclopedia.com/social-sciences/encyclopedias-almanacs-transcripts-and-maps/education-and-mobility}{www.encyclopedia.com/.../education-and-mobility}} unfair grading may perpetuate systemic discrimination and increase social inequality.

\vspace{-0.3em}
\paragraph{Dimension.} We can generally categorize written tests into those that test for content correctness (e.g., essay questions in a history test), and those that test for language skills (e.g., proper use of grammar). While there are tests that simultaneously assess both aspects, modern ATS systems often grade them separately \citep{ke2019automated}. We treat each aspect as a separate test here.

When grading students on content correctness, we would expect the ATS system to ignore linguistic variation and sensitive attributes as long as they do not affect the answer's validity. Hence, we would expect variation in these dimensions to have no effect on scores: answer length, language/vocabulary simplicity, alternative spellings/misspellings of non-keywords, grammatical variation, syntactic variation (especially those resembling transfer from a first language), and proxies for sensitive attributes. On the other hand, the system should be able to differentiate proper answers from those aimed at gaming the test \citep{chingaming2020,ding-etal-2020-dont}.

When grading students on language skills, however, we would expect ATS systems to be only sensitive to the relevant skill. For example, when assessing grammar use, we would expect the system to be sensitive to grammatical errors (from the perspective of the language variety the student is expected to use), but not to the other dimensions mentioned above (e.g., misspellings).

\vspace{-0.3em}
\paragraph{Actors.} Relevant experts include teachers of the subjects where the ATS systems will be deployed, linguists, and computer scientists. The stakeholders (students) may be represented by student unions (at the university level) or focus groups comprising a representative sample of the student population.

\section{Implications for Policy}\label{sec:policy}
\vspace{-0.3em}
There is a mounting effort to increase accountability and transparency around the development and use of NLP systems to prevent them from amplifying societal biases. DOCTOR is highly complementary to the model card approach increasingly adopted\footnote{\href{http://huggingface.co/models}{huggingface.co/models};\\\hspace*{1.75em}\href{https://github.com/ivylee/model-cards-and-datasheets}{github.com/ivylee/model-cards-and-datasheets};\\\hspace*{1.75em}\href{https://blog.einstein.ai/model-cards-for-ai-model-transparency}{blog.einstein.ai/model-cards-for-ai-model-transparency}} to surface oft hidden details about NLP models: Developers simply need to list the tested dimensions, metrics, and score on each dimension in the model card. Crucially, reliability tests can be used to highlight fairness issues in NLP systems by including sensitive attributes for the target population, but it is paramount these requirements reflect \emph{local concerns} rather than any prescriptivist perspective \citep{sambasivan2021re}.

At the same time, the ability to conduct quantitative, targeted reliability testing along specifiable dimensions paves the way for reliability standards to be established, with varying levels of stringency and rigor for different use cases and industries. We envision minimum safety and fairness standards being established for applications that are non-sensitive, not safety-critical, and used in unregulated industries, analogous to standards for household appliances. Naturally, applications at greater risks \citep{li2020toward} of causing harm upon failure should be held to stricter standards. Policymakers are starting to propose and implement regulations to enforce transparency and accountability in the use of AI systems. For example, the European Union’s General Data Protection Regulation grants data subjects the right to obtain ``meaningful information about the  logic involved'' in automated decision systems \citep{gdpr2016}. The EU is developing AI-specific regulation \citep{euaiwhitepaper2020}: e.g., requiring developers of high-risk AI systems to report their ``capabilities and limitations, ...\ [and] the conditions under which they can be expected to function as intended''. In the U.S., a proposed bill of the state of Washington will require public agencies to report ``any potential impacts of the automated decision system on civil rights and liberties and potential disparate impacts on marginalized communities'' before using automated decision systems \citep{sb51162021}.

One may note that language in the proposed regulation is intentionally vague. There are many ways to measure bias and fairness, depending on the type of model, context of use, and goal of the system. Today, companies developing AI systems employ the definitions they believe most reasonable (or perhaps easiest to implement), but regulation will need to be more specific for there to be meaningful compliance. DOCTOR's requirement to explicitly define specific dimensions instead of a vague notion of reliability will help policymakers in this regard, and can inform the ongoing development of national \citep{nistai2019} and international standards\footnote{\href{https://ethicsstandards.org/p7000}{ethicsstandards.org/p7000}}.

While external algorithm audits are becoming popular, testing remains a challenge since companies wishing to protect their intellectual property may be resistant to sharing their code \citep{johnsonauditing2021}, and implementing custom tests for each system is unscalable. Our approach to reliability testing offers a potential solution to this conundrum by treating NLP systems as black boxes. If reliability tests become a legal requirement, regulatory authorities will be able to mandate independently conducted reliability tests for transparency. Such standards, combined with certification programs (e.g., IEEE's Ethics Certification Program for Autonomous and Intelligent Systems\footnote{\href{https://standards.ieee.org/industry-connections/ecpais.html}{standards.ieee.org/industry-connections/ecpais.html}}), will further incentivize the development of responsible NLP, as the companies purchasing NLP systems will insist on certified systems to protect them from both legal and brand risk. To avoid confusion, we expect certification to occur for individual NLP systems (e.g., an end-to-end question answering system for customer enquiries), rather than for general purpose language models that will be further trained to perform some specific NLP task. While concrete standards and certification programs that can serve this purpose do not yet exist, we believe that they eventually will and hope our paper will inform their development. This multi-pronged approach can help to mitigate NLP's potential harms while increasing public trust in language technology.

\section{Challenges and Future Directions}
While DOCTOR is a useful starting point to implement reliability testing for NLP systems, we observe key challenges to its widespread adoption. First, identifying and prioritizing the dimensions that can attest a system's reliability and fairness. The former is relatively straightforward and can be achieved via collaboration with experts (e.g., as part of the U.S.\ NIST's future AI standards \citep{nistai2019}). The latter, however, is a question of values and power \citep{noble2018algorithms,mohamed2020decolonial,leins-etal-2020-give}, and should be addressed via a code of ethics and ensuring that all stakeholders are adequately represented at the decision table.

Second, our proposed method of reliability testing may suffer from similar issues plaguing automatic evaluation metrics for natural language generation \citep{novikova-etal-2017-need,reiter-2018-structured,kryscinski-etal-2019-neural}: due to the tests' synthetic nature they may not fully capture the nuances of reality. For example, if a test's objective were to test an NLP system's reliability when interacting with African American English (AAE) speakers, would it be possible to guarantee (in practice) that all generated examples fall within the distribution of AAE texts? Potential research directions would be to design adversary generation techniques that can offer such guarantees or incorporate human feedback \citep{nguyen2017reinforcement,kreutzer2018can,stiennon2020learning}.

\section{Conclusion}
Once language technologies leave the lab and start impacting real lives, concerns around safety, fairness, and accountability cease to be thought experiments. While it is clear that NLP can have a positive impact on our lives, from typing autocompletion to revitalizing endangered languages \citep{zhang-etal-2020-chren}, it also has the potential to perpetuate harmful stereotypes \citep{bolukbasi2016,sap-etal-2019-risk}, perform disproportionately poorly for underrepresented groups \citep{translation-arrest2017,bridgeman2012comparison}, and even erase already marginalized communities \citep{bender2021}.

Trust in our tools stems from an assurance that stakeholders will remain unharmed, even in the worst-case scenario. In many mature industries, this takes the form of reliability standards. However, for standards to be enacted and enforced, we must first operationalize ``reliability''. Hence, we argue for the need for reliability testing (especially worst-case testing) in NLP by contextualizing it among existing work on promoting accountability and improving generalization beyond the training distribution. Next, we showed how adversarial attacks can be reframed as worst-case tests. Finally, we proposed a possible paradigm, DOCTOR, for how reliability concerns can be realized as quantitative tests, and discussed how this framework can be used at different levels of organization or industry.

\section*{Acknowledgements}
Samson is supported by Salesforce and Singapore's Economic Development Board under the Industrial Postgraduate Programme. Araz is supported by the NUS Centre for Trusted Internet and Community through project CTIC-RP-20-02.

\section*{Broader Impact}
Much like how we expect to not be exposed to harmful electric shocks when using electrical appliances, we should expect some minimum levels of safety and fairness for the NLP systems we interact with in our everyday lives. As mentioned in \Cref{sec:intro}, \Cref{sec:doesnlp}, and \Cref{sec:policy}, standards and regulations for AI systems are in the process of being developed for this purpose, especially for applications deemed ``high-risk'', e.g., healthcare \citep{euaiwhitepaper2020}. Reliability testing, and our proposed framework, is one way to approach the problem of enacting enforceable standards and regulations.

However, the flip side of heavily regulating every single application of NLP is that it may slow down innovation. Therefore, it is important that the level of regulation for a particular application is proportionate to its potential for harm \citep{gdc2019}. Our framework can be adapted to different levels of risk by scaling down the implementation of some steps (e.g., the method and depth in which stakeholder consultation happens or the comprehensiveness of the set of testing dimensions) for low-risk applications.

Finally, it is important to ensure that any tests, standards, or regulations developed adequately represents the needs of the most vulnerable stakeholders, instead of constructing them in a prescriptivist manner \citep{hagerty2019global}. Hence, DOCTOR places a strong emphasis on involving stakeholder advocates and analyzing the impact of an application of NLP on the target community.

\bibliography{custom}
\bibliographystyle{acl_natbib}

\clearpage
\appendix
\section*{Appendix}
\vspace{-0.3em}
\section{Testing Dimensions: Detecting Violent Content on Social Media}\label{app:case}
\vspace{-0.3em}
In this second case study, we apply DOCTOR for measuring the reliability of a violent content detection system for English social media posts. Although we limit this discussion to the U.S., this is a growing global problem \citep{laubhate2019} that can lead to deadly outcomes \citep{rajagopalan2018}. In this hypothetical use case, the NLP system may automatically remove violent content or alert content moderators to potential violations of the social media company's acceptable use policy. Moderators can decide if specific content should be removed, and if necessary, notify law enforcement to avert pending violence (e.g., threats against individuals, planned violent events). As a result of the 1996 Communications Decency Act\footnote{\href{https://www.fcc.gov/general/telecommunications-act-1996}{fcc.gov/general/telecommunications-act-1996}}, social media platforms have broad latitude \citep{klonick2018new} to develop their own policies for acceptable content and how they handle it. In this scenario, the compliance officer of the company developing the system is responsible for making sure it does not discriminate against specific user demographics.

Research has shown that hate speech can lead to hateful actions \citep{marsters2019}. In many cases, individuals posted their intents online prior to committing violence \citep{cohen2014detecting}. When identifying content to remove and especially when involving law enforcement, it is important to distinguish between ``Hunters" --- those who act --- and ``Howlers" --- those who do not \citep{marsters2019}. This is to avoid wrongly detaining individuals who have no intention of committing violence, even if their words are indefensible. Between these extremes, posters may harass, stalk, dox, or otherwise abuse victims from a distance, therefore it is still necessary to flag, remove, and potentially track or document violent content.

\vspace{-0.3em}
\paragraph {Linguistic landscape.}
We focus solely on English speakers, but we acknowledge that the actual linguistic landscape is much more complex (over 350 languages). Posters on social media may speak English as their first language or as a second language and they often code-switch/-mix. Standard American English is used for business purposes in the U.S. but there are other frequently used language varieties including African American English (AAE), Cajun Vernacular English, and three different Latinx (Hispanic) vernacular Englishes.

\vspace{-0.3em}
\paragraph {Stakeholder Impact.}
The key stakeholders that will be impacted are those most often facing violent threats online: minorities, women, immigrants, and the LGBTQ community \citep{amnesty2018,ganeshhate2018,davidson-etal-2019-racial,wakefield2020}. Additionally, anyone that posts content on the social media site is a stakeholder. Unfortunately, the very communities that are often the target of violent posts are also  often wrongly flagged as posting toxic content themselves due to racial biases present in the training data \citep{sap-etal-2019-risk,davidson-etal-2019-racial}. Given the risk of harm to victims if the system misses violent posts from hunters or misidentifies legitimate content as violent and notifies law enforcement, it is critical the right balance of false positives and false negatives is achieved in flagging content.

\vspace{-0.3em}
\paragraph {Dimensions.}
There are two tasks under consideration here: identifying violent content and identifying Hunters who ``truly intend to use lethal violence'' \citep{marsters2019}. In the first task, the system is looking for content that negatively targets a socially defined group. Additionally, the content includes not only hate speech (e.g., profanity, epithets, vulgarity) but also content that incites others to hatred or violence. Since content written in AAE has been shown to be flagged as toxic more often \citep{sap-etal-2019-risk,davidson-etal-2019-racial}, we must ensure that the system is reliable when encountering dialectal variation. Additionally, due to the casual environment of social media, multilingual speakers often code-switch and code-mix. Hence, we expect variation in these dimensions to have no effect on the system's predictions: alternative spellings, morphosyntactic variation, word choice, code-mixing, idioms, and references to and manifestations of sensitive attributes and their proxies. However, we must expect the system to be sensitive to in-group and out-group usage of reclaimed slurs so that the in-group usage does not result in a flag while out-group usage result in flagged posts.

When identifying hunters, we may expect the system to be sensitive to uses of first person pronouns, certainty adverbs, negative evaluative adjectives, and modifiers \citep{marsters2019}. However, in order to avoid unfairly penalizing vernacular English speakers we should expect the system's predictions to be equally unaffected by variation in the dimensions listed for the first task.

\begin{table*}[p]
\small
    \centering
    \begin{tabular}{l|c|c}
    \toprule
 & \multirow{15}{*}{Orthography} & Hyphenation\\
 & & Capitalization\\
& & Punctuation\\
& & Reduplication of letters\\
& & Emojis/emoticons \\
& & Homonyms \\
& & Disemvoweling \citep{eger-benz-2020-hero}\\
& & Homophones (e.g., accept vs.\ except) \citep{eger-benz-2020-hero} \\
& & Accidental misspellings \citep{belinkov2018synthetic}\\
& & Intentional alternative spellings (e.g., Yas, thru, startin)\\
& & Open compound concatenation (e.g., couch potato/couchpotato) \\
& & Dialectal differences (e.g., favor vs.\ favour) \citep{SinghGR18}\\
& & Mixing writing scripts \citep{tan-adv-polyglots21}\\
& & Transliteration \\
\cmidrule{2-3}
 & \multirow{4}{*}{Morphology} & Grammatical gender shifts \\
& & Grammatical category \citep{tan-etal-2020-morphin} \\
& & Dialectal differences \citep{tan-etal-2020-morphin}\\
Linguistic & & Clitics  \\
\cmidrule{2-3}
Phenomena & \multirow{5}{*}{Lexicon} & Dialectal variation (e.g., fries vs.\ chips)\\
& & Synonyms/Sememes \citep{zang2020word} \\
& & Vocabulary simplicity/complexity \\
& & Cross-lingual synonyms \citep{tan-adv-polyglots21}\\
& & Loanwords\\
\cmidrule{2-3}
& Semantics & Idioms (e.g., finer than frog hair) \\
\cmidrule{2-3}
& \multirow{6}{*}{Syntax} & Matching number and tense\\
& & Word/phrase order (especially for languages without strict word ordering)\\
&& Prepositional variation (e.g., stand on line vs. stand in line) \\
&& Syntactic variation \citep{iyyer-etal-2018-adversarial}\\
& & Sentence simplicity/complexity \\
&& Code-mixing \citep{tan-adv-polyglots21}\\
\cmidrule{2-3}
& & Register (e.g., formality)\\
& Discourse  & Conversational style (involvement/considerateness) \citep{tannen2005conversational}\\
& \&  & Discourse markers / connector words \\
& Pragmatics & Cross-cultural differences \\
& & Code-switching \\
\midrule
\multirow{20}{*}{Sensitive Attributes} & \multirow{4}{*}{Gender Identity} & Gender pronouns \\
& & Names \\
& & Reclaimed slurs \\
& & Genderlects \citep{tannen1991you,dunn2014gender}\\
\cmidrule{2-3}
& \multirow{3}{*}{Race} & Names \\
& & Reclaimed slurs \\
& & Race-aligned language varieties\\
\cmidrule{2-3}
& Age & Age/generation-aligned language styles \citep{hovy-etal-2020-sound}\\
\cmidrule{2-3}
& \multirow{2}{*}{Religion} & Names \\
& & Reclaimed slurs \\
\cmidrule{2-3}
& Sexual Orientation & Reclaimed slurs \\
\cmidrule{2-3}
& Disability status & Associated adjectives \citep{hutchinson-etal-2020-social}\\
\cmidrule{2-3}
& \multirow{2}{*}{Place of origin} & Location names (e.g., cities, countries)\\
& & Figures of speech \\
\cmidrule{2-3}
& Proxies & Geographic locations (for ethnicity, socioeconomic status)\\
\midrule
\multirow{5}{*}{Malicious Attacks} & \multirow{2}{*}{Black-box} & Rule-based \citep{alzantot-etal-2018-generating,jin2019bert}\\
& & Model-based \citep{garg2020bae,li2020bert}\\
\cmidrule{2-3}
& Gradient-based & HotFlip \citep{ebrahimi-etal-2018-hotflip}, Universal Triggers \citep{wallace-etal-2019-universal}\\
\cmidrule{2-3}
& Policy-based & Adversarial negotiation agent \citep{cheng-etal-2019-evaluating}\\
\bottomrule
    \end{tabular}
    \vspace{-0.5em}
    \caption{Taxonomy of possible dimensions with references to linguistics literature and existing adversarial attacks that could be used as worst-case tests. Linguists are best equipped to decide which linguistic phenomena are high priority for each use case, ethicists for sensitive attributes, and NLP practitioners for malicious attacks.}
    \label{tab:taxonomy}
\end{table*}

\end{document}

%% file: math_commands.tex
\usepackage{amsfonts,bm}

\newcommand{\mytilde}{{\raise.17ex\hbox{$\scriptstyle\mathtt{\sim}$}}}

\def\eqref#1{equation~\ref{#1}}

\def\1{\bm{1}}

\DeclareMathAlphabet{\mathsfit}{\encodingdefault}{\sfdefault}{m}{sl}
\SetMathAlphabet{\mathsfit}{bold}{\encodingdefault}{\sfdefault}{bx}{n}

\def\gD{{\mathcal{D}}}

\def\gM{{\mathcal{M}}}

\def\gS{{\mathcal{S}}}

\def\gX{{\mathcal{X}}}
\def\gY{{\mathcal{Y}}}

\DeclareMathOperator*{\argmin}{arg\,min}

\crefformat{section}{\S#2#1#3} 
\crefname{algorithm}{Alg.}{Algs.}
\crefformat{subsection}{\S#2#1#3}
\Crefname{equation}{Eq.}{Eqs.}
\Crefname{figure}{Fig.}{Figs.}

\algnewcommand\algorithmicswitch{\textbf{switch}}
\algnewcommand\algorithmiccase{\textbf{case}}
\algdef{SE}[SWITCH]{Switch}{EndSwitch}[1]{\algorithmicswitch\ #1\ \algorithmicdo}{\algorithmicend\ \algorithmicswitch}%
\algdef{SE}[CASE]{Case}{EndCase}[1]{\algorithmiccase\ #1}{\algorithmicend\ \algorithmiccase}%
\algtext*{EndSwitch}%
\algtext*{EndCase}%